\documentclass[]{interact}

\usepackage{epstopdf}
\usepackage[caption=false]{subfig}

\usepackage[natbibapa,nodoi]{apacite}
\renewcommand\bibliographytypesize{\fontsize{10}{12}\selectfont}

\usepackage{todonotes}
\usepackage{indentfirst}
\usepackage{setspace}
\usepackage{soul}
\usepackage{float}
\usepackage[colorlinks=true, linkcolor=black, urlcolor=blue, citecolor=blue, anchorcolor=blue]{hyperref}
\usepackage{wrapfig}
\usepackage{comment}
\usepackage{textgreek}
\usepackage{url}
\usepackage[font=footnotesize,labelfont=it]{caption}

\usepackage{enumitem, comment, xifthen}
\usepackage{amsmath,amssymb,amsthm, amsfonts}
\usepackage[T1]{fontenc}
\usepackage{makecell}
\renewcommand*{\bibfont}{\small}

\usepackage{booktabs}
\usepackage{hhline}
\usepackage{array,multirow}
\usepackage{lipsum}
\usepackage{siunitx,etoolbox}

\usepackage{graphicx}

\usepackage{tikz}
\usepackage{lipsum}

\theoremstyle{plain}

\theoremstyle{definition}

\theoremstyle{remark}

\newcommand{\kp}[1]{{\color{black}{#1}}}

\begin{document}
	
	
	\title{Explainable prediction of Qcodes for NOTAMs\\ using column generation}
	
	\author{
		\name{Krunal Kishor Patel\textsuperscript{a}, Guy Desaulniers\textsuperscript{b}\thanks{CONTACT Guy Desaulniers. Email: guy.desaulniers@gerad.ca}, Andrea Lodi\textsuperscript{a,c} and Freddy Lecue\textsuperscript{d}}
		\affil{\textsuperscript{a}CERC, Polytechnique Montr\'eal, Montr\'eal, Canada; \textsuperscript{b}Polytechnique Montr\'eal and GERAD, Montr\'eal, Canada; \textsuperscript{c}Jacobs Technion-Cornell Institute, Cornell Tech and Technion - IIT, New York, USA; \textsuperscript{d}Inria, Sophia Antipolis, France}
	}
	
	\maketitle
	
	\begin{abstract}
		A NOtice To AirMen (NOTAM) contains important flight route related information. To search and filter them, NOTAMs are grouped into categories called QCodes. In this paper, we develop a tool to predict, with some explanations, a Qcode for a NOTAM. We present a way to extend the interpretable binary classification using column generation proposed in \cite{Dash2018} to a multiclass text classification method. We describe the techniques used to tackle the issues related to one-vs-rest classification, such as multiple outputs and class imbalances. Furthermore, we introduce some heuristics, including the use of a CP-SAT solver for the subproblems, to reduce the training time. Finally, we show that our approach compares favorably with state-of-the-art machine learning algorithms like Linear SVM and small neural networks while adding the needed interpretability component.
	\end{abstract}
	

	\section{Introduction}
	\label{intro}
	
	A NOtice To AirMen or NOtice To Air Missions (NOTAM) \citep{notamFaa} is a notice sent to alert aircraft pilots of potential hazards along a flight route or at a location that could affect the flight. Generally, a NOTAM states the abnormal status of a component of the National Airspace System (NAS). NOTAMs concern the establishment, condition, or change of any facility, service, procedure, or hazard in the NAS. Sometimes a NOTAM is not alerting of potential hazard, but only providing some practical information that is not critical.
	
	The NOTAMs are grouped by a label called Qcode \citep{notamQcodesFaa}. The aircraft pilots use this Qcode to search and filter the NOTAMs. Each NOTAM is assigned one Qcode. The Qcode is a five letter label. The first letter is always `Q'. The second and the third letter identify the subject being reported. The fourth and fifth letters identify the status of operation of the subject being reported. These Qcodes are assigned manually to each message by the airport authorities writing NOTAMs. Since the Qcode categories are too many and partially redundant, people sometimes assign them wrongly, or do not assign any and choose “QXXXX”. Our goal is to predict the Qcode for a NOTAM given its message content. 
	
	\kp{Standard machine learning algorithms can predict the Qcode for a given message. However, they lack explanation. Using boolean decision rules, we are able to tackle the explainability part. But, to achieve an accuracy similar to that obtained by standard machine learning algorithms, we have to predict a small list of Qcodes instead of a single Qcode for a given message. A human will then pick a final Qcode from this list (along with the explanations). We believe that this is better than picking a Qcode from a large pool of possible Qcodes or blindly trusting a machine learning model's output without any explanation.}
	
	Note that a NOTAM contains many fields other than just the message. However, \kp{the industrial partner Thales asked for a solution using} only the message field to predict the Qcode.
	
\begin{figure}[H]
		\centering
		\includegraphics[width=\textwidth]{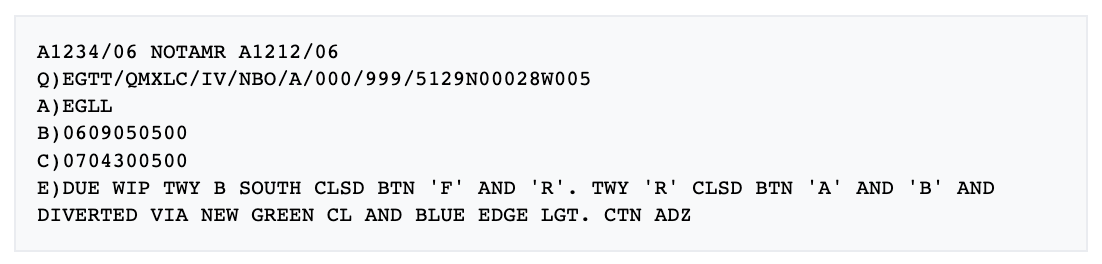}
		\caption{A typical NOTAM for London Heathrow Airport.}
		\label{fig:notam_example}
	\end{figure}

	Figure \ref{fig:notam_example} shows a typical NOTAM example for London Heathrow Airport.
	
	As shown, the NOTAM has many fields. We primarily focus on the Qcode (QMXLC) and the message (field E) that reads as follows: 
	
	\begin{quote}
		
		Due to work in progress (DUE WIP), taxiway "B South" is closed between "F" and "R" (TWY B SOUTH CLSD BTN 'F' AND 'R'). Taxiway "R" is closed between "A" and "B" (TWY 'R' CLSD BTN 'A' AND 'B') and is diverted via a new green centre line and blue edge lighting (AND DIVERTED VIA NEW GREEN CL AND BLUE EDGE LGT). Caution advised (CTN ADZ).
	\end{quote}
	
	
	Explainability is a well-recognized goal for machine learning prediction tasks. The Qcodes can be predicted without explanation for the NOTAMs. However, since the Qcodes are assigned by humans (the airport authorities) and used by humans (the pilots), the explanation for each Qcode increases confidence and ensure trust by airport authorities. This is what we are looking to achieve in this work.
	
	
	
	
	To classify NOTAMs into Qcodes, we use an extension of the mixed-integer programming (MIP) model proposed in \cite{Dash2018} and solved by column generation. This approach is a binary classification method that assumes that the inputs have binary features. The output is a boolean decision rule that evaluates to TRUE if the input is classified as positive and FALSE otherwise. More precisely, the boolean decision rule is a Disjunctive Normal Form (DNF), OR-of-ANDs formula. A DNF formula is a set of clauses. If any of the clause evaluates to TRUE, the formula evaluates to TRUE. Each clause contains a set of binary features and is set to TRUE if all the features are TRUE in the clause. An example of a DNF rule with two clauses is "($x_1$ AND $x_2$) OR ($x_3$ AND $x_5$ AND $x_7$)", where $x_i$, $i=1,2,3,5,7$, denote a binary feature. These binary rules also serve as explanations for the output class.
	
	In order to extend the approach in \cite{Dash2018} to our context, we had to face a few challenges. First, the NOTAM classification task is a multiclass classification problem. A one-vs-rest classification method is a natural way to extend a binary classification method to a multiclass classification context. In addition, the model proposed in \cite{Dash2018} is designed for balanced datasets (similar number of positive and negative inputs in the training data). However, because we are using one-vs-rest classification, the number of negative data points is much larger compared to the number of positive data points. Another issue is that an input can satisfy the boolean decision rules of more than one class. This is a common issue with one-vs-rest methods. Finally, in our context, we could not use the entire training dataset for learning the boolean decision rule for a class because that leads to very large linear programs. In this paper, we describe how to effectively adapt and extend the method introduced in \cite{Dash2018} to address all these challenges, and we show that the resulting extension is a very competitive approach to provide explainable classifications of NOTAMs. 
	
	
	
	
	
	The paper is organized as follows. In Section \ref{sec:rule-gen}, we briefly discuss boolean decision rules generation and provide an overview of binary classification using the boolean decision rules framework of \cite{Dash2018}. In Section \ref{sec:Multiclass}, we describe how we extend the binary classification framework to a multiclass classification framework. In Section \ref{sec:computation}, we describe the computational approach for the classification, while in Section \ref{sec:largedata}, we present the evaluation of our approach for the NOTAM classification problem. Finally, Section \ref{sec:conclusion} draws some conclusions and describe\kp{s} future research objectives.
	
	\section{Overview of Boolean Decision Rules \kp{and other classification methods}}
	\label{sec:rule-gen}
	
	\kp{Boolean decision rules are easy to interpret. The satisfied clauses can be used to explain a prediction. For multiclass classification, decision trees provide the most interpretable classification. In fact, to the best of our knowledge, there are no other multiclass classification methods that are interpretable or explainable. So, besides comparing the approach that we propose to decision trees, we also benchmark it against some standard, but non-explainable, machine learning algorithms for multiclass classification: Bernoulli Naive Bayes, Neural network, XGBoost \citep{chen2016}, Linear Support Vector Machine \citep{cortes1995support}. Among these algorithms, only Support Vector Machine is a binary classification method and is extended to a multiclass classification method using the one-vs-rest approach. Hence, we survey some of the alternative ways to generate boolean decision rules and describe the approach used in \cite{Dash2018} after that. Our extension to multiclass classification can be used for all of them since they all output a boolean decision rule.}
	
	Boolean decision rules can be generated in many ways. We selected the approach in \cite{Dash2018} because it focuses both on the rule complexity and the accuracy, thus prioritizing interpretability. Some of the alternatives for learning boolean decision rules are as follows. The AQ method \citep{cervone2010algorithm} attempts to learn the rule by considering one clause at a time heuristically. The clauses only cover the positive examples. The AQ method does not work if the dataset has noise (examples with same features belonging to different classes). The CN2 method \citep{clark1989cn2, clark1991rule} extends the AQ algorithm and also generates clauses that may cover a limited number of negative examples. It generates the clause leveraging ideas from the ID3 algorithm \citep{quinlan1986induction}. More precisely, ID3 is used for selecting decision tests in a node of a decision tree. The ID3 algorithm measures the impurity of a node in the decision tree by an entropy function. The decision test that minimizes the impurity is selected for each node. Similarly, the CN2 algorithm computes the impurity of the clauses (using the entropy function) to evaluate their quality. The clauses that cover mostly the positive examples and a few negative examples are less impure. Such clauses are added to the decision rules. This allows the CN2 algorithm to work with the data with noise. The RIPPER algorithm \citep{cohen1995fast} learns the rules by using a separate-and-conquer strategy. It starts by treating the least common class in the dataset. It generates the clauses by using a general-to-specific method for growing clauses. Then, it prunes the generated clauses depending on their performance in the validation set. These algorithms mostly aim for accuracy and do not focus on reducing the rule complexity. Hamming clustering \citep{muselli2002binary} is another way of learning boolean rules. Here, the rules are learned by clustering the inputs and then pruning to generalize. A detailed survey for rule learning is given in \cite{furnkranz2012foundations}.
	
	In \cite{Dash2018}, the authors describe a simple MIP model to construct a DNF boolean decision rule. In this model, they use binary variables to represent clauses. The clause is included in the decision rule if the corresponding binary variable takes value 1. We also want our decision rule to be interpretable. A decision rule that contains too many clauses or too large clauses is less interpretable. To avoid generating such rules, they use a hyperparameter $C$ for limiting the complexity of the decision rule. The complete model is based on the notation presented in Table \ref{tab:parameters}.
	
	\begin{table}[h!]
		\tbl{Model description: sets, parameters, and variables.}{
			\begin{tabular}{ll}
				\textbf{Sets} & \\
				$\mathcal{P}$ & Set of all positive examples in the dataset. \\
				$\mathcal{Z}$ & Set of all negative examples in the dataset. \\
				$\mathcal{K}$ & Set of all clauses. \\
				$\mathcal{K}_i$ & Set of all clauses satisfied by example $i \in \mathcal{P} \cup \mathcal{Z}$. \\
				
				\textbf{Parameters} & \\
				$c_k$& Complexity of clause $k \in \mathcal{K}$. Typically, it is set to 1 plus the length of the clause.\\
				$C$& Complexity of the explanation.\\
				
				\textbf{Decision Variables}& \\
				$w_k$& Binary variable indicating if clause $k \in \mathcal{K}$ is selected.\\
				$\xi_i$& Binary variable indicating if example $i  \in \mathcal{P}$ does not satisfy any selected clauses.
		\end{tabular}}
		\label{tab:parameters}
	\end{table}
	The resulting MIP reads as follows:
	\begin{subequations}
		\label{origmodel}
		\begin{align}
			z_{MIP} = \min \quad&  \sum_{i \in \mathcal{P}} \xi_i + \sum_{i \in \mathcal{Z}} \sum_{k \in \mathcal{K}_i} w_k \label{mas_obj} \\ 
			s.t. \quad & \xi_i + \sum_{k \in \mathcal{K}_i} w_k \geq 1, \quad \forall i \in \mathcal{P} \label{clause_sat}\\
			& \sum_{k \in \mathcal{K}} c_kw_k \leq C \label{clause_comp}\\
			&  \xi_i \geq 0, \quad \forall i \in \mathcal{P}\\ 
			& w_k \in \{0,1\}, \quad \forall k \in \mathcal{K}. \label{upper_bound}
		\end{align}
	\end{subequations}
	
	The objective function \eqref{mas_obj} minimizes the sum of the number of positive examples that do not satisfy any selected clauses (first term) and the number of positively-classified negative examples (second term).
	Constraints (\ref{clause_sat}) ensure that each positive example $i$ satisfies at least one of the selected clauses. If not, the corresponding penalty variable $\xi_i$ is set to 1. Indeed, the $\xi$ variables are implied binary, so they do not need to be recorded as binary variables. The constraint  (\ref{clause_comp}) limits the overall complexity of the generated decision rule, where the individual clause complexity is set to $c_k$. Finally, note that the objective minimizes a total penalty, where the penalty for a positive example is given by its corresponding $\xi$ variable and that for a negative example is set to the total number of selected clauses satisfied by that example. In a column generation context, we refer to model (\ref{origmodel}) as the integer master problem (integer MP) and its linear relaxation as the master problem (MP).
	
	Notice that in the MP, we have a variable for each clause. The number of possible clauses is exponential in the number of features. For this reason, we cannot solve the MP as it is. In \cite{Dash2018}, the authors use column generation for solving the MP. In this process \citep[see, e.g.,][]{Desrosiers2005}, the MP is initialized by the $\xi$ variables and a small number of $w$ variables. We refer to this linear program as a restricted master problem (RMP). Then, the $w$ variables are generated (priced) by solving a subproblem (SP) and added to the RMP as needed, i.e., when they have a negative reduced cost with respect to the dual solution of the current RMP. 
	
	More precisely, the reduced cost of column $w_k$ is given by $$\sum_{i \in \mathcal{Z}} \delta_i - \sum_{i \in \mathcal{P}} \mu_i \delta_i - \lambda c_k,$$ where $\mu_i \geq 0$ is the dual variable of the $i^{th}$ constraint (\ref{clause_sat}), $\lambda \leq 0$ is the dual variable of constraint (\ref{clause_comp}), and the binary variable $\delta_i$ is set to 1 if the example $i\in \mathcal{P} \cup \mathcal{Z}$ satisfies the clause $w_k$. 
	Using the notation provided in Table \ref{tab:sp_param}, the SP used to generate clause variables $w_k$ can be formulated as follows for a given RMP optimal dual solution $(\lambda^*,\mu^*)$: 
	
	\begin{table}[h!]
		\tbl{Subproblem description: sets, parameters, and variables.}
		{\begin{tabular}{ll}
				\textbf{Sets} & \\
				$J$ &  Set of all features.\\
				$S_i = \{j \in J: x_{ij} = 0\}$ & The zero-valued features in sample $i \in \mathcal{P} \cup \mathcal{Z}.$\\
				
				\textbf{Parameters} & \\
				$D$& Maximum size of the generated clause.\\
				
				\textbf{Decision Variables}& \\
				$z_j$& Binary variable indicating if feature $j$ is present in the generated clause.\\
				$\delta_i$& Binary variable indicating if example $i$ is satisfied by the generated clause.
		\end{tabular}}
		\label{tab:sp_param}
	\end{table}
	
	\begin{subequations}
		\label{origsp}
		\begin{align}
			z_{CG} = \min \quad& \sum_{i \in \mathcal{Z}} \delta_i - \lambda^* \left(1 + \sum_{j \in J} z_j \right)- \sum_{i \in \mathcal{P}} \mu_i^* \delta_i   \label{sp_obj} \\ 
			s.t. \quad & \delta_i + z_j \leq 1, \quad \forall j \in S_i, i \in \mathcal{P} \label{sp_one}\\
			& \delta_i \geq 1 - \sum_{j \in S_i}z_j, \quad \forall i \in \mathcal{Z} \label{sp_two}\\
			& \sum_{j \in J} z_j \leq D \label{sp_three}\\
			& z_j \in \{0,1\}, \quad \forall j \in J \label{sp_four} \\
			& \delta_i \geq 0, \quad \forall i \in \mathcal{P} \cup \mathcal{Z}.
		\end{align}
	\end{subequations}
	
	In the objective (\ref{sp_obj}), the term $1 + \sum_{j\in J} z_j$ is equal to the generated clause complexity, i.e., to 1 plus the clause length. This objective function corresponds to the reduced cost of the clause generated. A negative objective value identifies a clause with a negative reduced cost, which can be added to the RMP. The clause is constructed by selecting the features for which the corresponding $z$ variables are set to 1. The constraints (\ref{sp_one}) – (\ref{sp_two}) relate the selected features with the $\delta$ variables. Constraint (\ref{sp_three}) controls the size of the clause being generated. Here another hyperparameter $D$ is used.  Clearly, the value of $D$ should be less than or equal to the value of $C$. Note that the variables $\delta_i$ are implied binary, so they need not be encoded as binary. Indeed, constraints \eqref{sp_one} (resp. \eqref{sp_two}) are valid for all examples, but they become relevant only for the positive (resp. negative) ones because of the sign in the objective function.
	
	Column generation stops when no more negative reduced cost columns can be generated or when a maximum number of column generation iterations is reached. In the former case, the cost of the last RMP solution provides a lower bound on the optimal value of the integer MP. In the latter case, we do not have this guarantee. For solving the integer MP, we can use branch-and-price that embeds column generation within a branch-and-bound search tree \citep[see][]{Barnhart1998,Desaulniers2005}. Instead, we just solve the integer version of the final RMP using a commercial MIP solver, yielding a so-called restricted master heuristic \citep{JoncourEtAl2010}. 
	
	Overall, the process is as follows. We start with a training dataset. The integer MP (\ref{origmodel}) is then solved as described above. The computed solution gives us a DNF decision rule. We use this decision rule to classify new unseen inputs. If an input is classified as positive, the clauses that are satisfied by that example in the DNF rule serve as an explanation for “why the input was classified as positive?”.
	
	\section{Extending the model for multiclass classification}
	\label{sec:Multiclass}
	
	In this section, we extend the technique described in \cite{Dash2018} for explainable binary classification to our multiclass classification problem. Multiclass classification can be done by extending the binary classification framework to one-vs-rest classification. In the one-vs-rest classification, we train the binary classification model for each class in the input dataset. For each model, we use the training data inputs belonging to other classes as negative examples and inputs belonging to the class being trained as positive examples. 
	
	So, the overall training process is as follows. Consider a Qcode, we mark the inputs as positive or negative depending on whether they have that Qcode or not. Then, we solve the integer MP (\ref{origmodel}) using column generation and the restricted master heuristic. This procedure outputs a DNF rule as our explanation for that Qcode. We repeat this process for each Qcode and store the explanations. This ends our training phase. For evaluation, we consider an unseen input (message). We evaluate this message on each of the stored explanations (DNF rules). Then, we output a list of Qcodes for which the DNF rules evaluated to TRUE on this input. This list is called the candidate list. 
	
	\subsection{Change in the objective function}
	
	One of the major challenges is about handling the imbalance in the training data for each class. Note that the model above is designed to optimize accuracy. In our task, for each Qcode, there are very few positive examples compared to the number of negative examples. Hence, optimizing for accuracy leads to many false negatives, which is not desirable. Even if we limit the number of negative examples to match the number of positive examples, we cannot directly tune the recall (number of identified positives divided by total number of positives). Furthermore, using very few negative examples leads to a poor formula that does not generalize well for the unseen examples. One way to address this issue is to introduce a new hyperparameter $P$ that controls the recall during training. We change the objective function (\ref{mas_obj}) to have increased penalty for the false negatives by increasing the coefficient of each $\xi_i$ variable to $P$. The new objective is given by
	
	\begin{equation}
		z_{MIP} = \min \quad P\sum_{i \in \mathcal{P}} \xi_i + \sum_{i \in \mathcal{Z}} \sum_{k \in \mathcal{K}_i} w_k. \label{mas_obj_updated} 
	\end{equation}
	
	\subsection{Candidate weights}\label{sec:weights}
	
	Another of the challenges with the one-vs-rest classification method is picking a winner when more than one class shows positive results for the membership. This may result in a long candidate list. This is a common challenge across each classifier that extends the binary classification to multiclass classification using one-vs-rest methods. Linear support vector machines (SVM) is a good example for this. Linear SVM is a binary classification method. In the Scikit-learn \citep{sklearn_api} linear SVM implementation, the winner is picked by using the distance between the data point and the decision boundary (a hypersurface that separates the two classes). For linear SVM, the decision boundary is a line separating the two classes. The distance between a data point and the decision boundary indicates how certain the model is about the prediction. When more than one class shows positive results, the class, for which the distance to the decision boundary is maximized, is picked as winner. 
	
	If we train for a small number of Qcodes (say, up to 10), then the mean candidate list length is not large. However, we face the issue of large candidate list lengths when we train for a large number of Qcodes (at least 100). This motivates us to eliminate some of the redundant candidates from the list. Unlike SVM, we do not have a distance function because our decision boundary is a discrete function. However, we can assign some weight to each candidate in the list in a different way such that a higher weight corresponds to a higher likelihood of the example belonging to that class. We then sort the candidates in decreasing order of their weights and keep only the first few candidates with the highest weights. The candidate weights are computed as follows. 
	
	For each Qcode, we compute the explanation by solving the integer MP (\ref{origmodel}). Recall that each explanation is an OR of clauses (DNF). Now consider a clause in the explanation. We want to assign a weight to this clause so that it reflects how strongly it indicates whether the example belongs to that candidate or not. We use two numbers to compute the weight for a clause, namely 
	\begin{enumerate}
		\item Positive accuracy $W_p$: Number of positive training examples that satisfy the clause divided by the total number of positive examples in the training data. 
		\item Negative accuracy $W_n$: Number of negative training examples that satisfy the clause divided by the total number of negative examples in the training data.
	\end{enumerate}
	Naturally, we want to assign a high weight to a clause for which $W_p$ is large and $W_n$ is small. We tried a few different ways for computing the weights combining $W_p$ and $W_n$, and ultimately, we have chosen to set them as $W_{diff} = W_p - W_n$. \kp{This combination is simple and works better than some other combinations like $W_{ratio} = W_p / (W_n + \epsilon)$, where $\epsilon$ is a small positive number. These are the reasons we have selected it.}
	
	In our data, we do not have the same number of positive examples for each label. A clause that is generated with a small amount of training data gives us less confidence compared to a clause that is generated using a large amount of training data. So, to factor that, we also multiply the weights by the number of total training data used for that Qcode denoted as $n_k$, thus finally giving
	$W_{diff}^* = (W_p - W_n) \times n_k$.
	
	Consider a new unseen message and a candidate list. We want to compute the weight for each Qcode in this list. For a Qcode, we know the weights of each clause in the explanation and we only consider the clauses that are satisfied by this new example (there is at least one such clause if the Qcode is in the candidate list). The weight of the Qcode is computed as the sum of the weights of these clauses.  
	
	\section{Computational approach}
	\label{sec:computation}
	
	This section presents the computational approach for predicting the Qcodes of NOTAM messages. We start by describing the preprocessing of the data in Section \ref{sec:preprocess}. Then, in Section \ref{sec:solving}, we discuss the training part, which involves solving a sequence of integer MPs (\ref{origmodel}), namely, one per Qcode. Finally, we describe in Section \ref{sec:output} how prediction is performed for new unseen NOTAMs.
	
	\subsection{Data Preprocessing}
	\label{sec:preprocess}
	
	We first preprocess the data to make it ready for the training task. This requires mainly three tasks: Cleaning, Splitting and Binarization. For this task, our industrial partner Thales gave us a sample dataset for training. The dataset has a total of 21,358 NOTAMs and a total of 603 different Qcodes. \kp{It is available at \url{https://github.com/krooonal/NOTAM_explainable_prediction_data}.}
	
	\textbf{Cleaning:} First, we discard all the Qcodes for which we do not have a sufficient number of positive examples to avoid poor generalization. We keep only the Qcodes for which there are at least 10 positive examples in the dataset. After removing the Qcodes with less than 10 positive examples, we are left with 20,223 NOTAMs and 204 Qcodes. This step eliminates a lot of Qcodes, but the same approach can be extended to generate explanations for the eliminated Qcodes, if we get more data. Moreover, for this prediction task, we only use the message field of the NOTAMs as input and Qcode field as output. We discard all the other information.
	
	In the NOTAMs, we remove all the punctuation characters except '/' (since it is very frequent) and replace them by a space. All the messages are already in capital letters. No other preprocessing is done on the input.
	
	\textbf{Splitting:} We split the data into three sets: Training (60\%), Validation (20\%), and Test (20\%). Training data is used for training, Validation data is used for tuning the hyperparameters and Test data is used for estimating the performance of the model.
	
	\textbf{Binarization:} The MP (\ref{origmodel}) requires the data to have binary features. So, we convert the data into a \textit{binary bag of words} representation as follows. We iterate through the Training dataset and select the most frequent 1,000 words. The number of words is a hyperparameter and  can be further tuned. However, we observe that using more words does not justify the improvement compared to the increase in training time for our data. Those most frequent words are our binary features for our input:  for each word, we have an input binary entry taking value 1 if the input NOTAM contains that word and 0 otherwise. Note that with this representation, we lose the information about the order in which the words appear. This can be an issue for many text classification applications, but in our case the messages are short and simple, so the prediction with high accuracy can still be done with this representation.
	
	\subsection{Training}
	\label{sec:solving}
	
	To determine the DNF rule for each Qcode, the integer MP (\ref{origmodel}) is solved considering the training dataset. To do so, we use column generation and the restricted master heuristic as described in Section~\ref{sec:rule-gen}, with the following specificities. 
	
	
	Column generation stops when no more negative reduced cost column can be found or when a maximum number of column generation iterations $I$ is reached, where one iteration corresponds to solving the RMP and the SP once each. $I$ is a hyperparameter. 
	
	Since the training data is large, we do not include all the negative examples while training for a particular Qcode. We randomly select a fraction of negative examples as $N (\geq 1)$ times the number of positive examples. Hence, we also have $N$ as a hyperparameter. 
	
	In the SP \eqref{origsp}, $D$ is a hyperparameter that limits the size of the clause generated. The value of $D$ should be clearly less than the value of $C$ in the integer MP (\ref{origmodel}). When  generating clauses with $D=C$, we find that the generated clause size never exceeds 3. So, we use $D = 3$. Choosing a small value for $D$ helps to reduce the training time.
	
	In terms of solving time, as shown in Table \ref{tab:training-stats}, the real bottleneck is associated with solving SP (\ref{origsp}). To reduce the training time, we use a heuristic to find negative reduced cost clauses. For each Qcode, we start by selecting at most 200 of the most frequent words in the positive examples. Then, we remove all the words that appear in less than 2\% of the examples. Using this small subset of words, we compute all size 1 and size 2 clauses and their reduced costs. We add all the clauses with negative reduced costs to the RMP. We only solve the SP \eqref{origsp} if we fail to find any clause with negative reduced cost with this heuristic. 
	
	For solving SP (\ref{origsp}), we observed that the OR-tools (version 9.2) solver CP-SAT \citep{Google-OR-Tools}, a constraint programming solver using a SAT solver as back-end for tree search, is approximately 20\% faster than Gurobi \citep{gurobi} (version 9.1.1). This is probably because of the combinatorial structure of the SP. For this reason, the solving approach in CP-SAT that uses significantly less LP solving is beneficial. However, CP-SAT can only accept integer coefficients. So, to use it, we have to round the dual values to integer values: we round all the coefficients in the objective (\ref{sp_obj}) up (i.e., all the dual variables $\lambda^*$ and $\mu^*$ are rounded down). So, the columns generated with CP-SAT (the ones with negative reduced cost) are always valid. To minimize the error due to rounding, we scale the objective coefficients by multiplying them by 100. In addition, \kp{to avoid overfitting, }we only add the columns that have reduced cost less than $-10^{-2}$. 
	To achieve a faster convergence, we add all the negative reduced cost columns we find during the solving process, as this number of columns is currently not critical in our application. 
	
	Notice that while using CP-SAT, we explicitly require all implicitly-binary $\delta$ variables to be binary. Since solving SP is highly time-consuming, we use 8 threads with the CP-SAT solver, which is configured with the default parameter values.
	
	To further reduce the search space in SP \eqref{origsp}, we fix to 0 all the $z$ variables that correspond to words that do not appear in any of the positive examples. This removes many feasible solutions of SP \eqref{origsp}, but preserves all the solutions that correspond to negative reduced cost clauses. This is because only the last term in the SP objective \eqref{sp_obj} has negative coefficients. So, at least one positive example has to satisfy the generated clause in order to have a negative reduced cost. 


	For solving the RMP at each column generation iteration, we use GLOP, an open-sourced linear programming solver in Google OR-tools, version 9.2 \citep{Google-OR-Tools}, with a single thread. For solving the final RMP with integrality requirements, we use Gurobi, version 9.1.1 \citep{gurobi}. 
	
	For all the training process, we used a CPU server with 2 sockets of Intel(R) Xeon(R) Gold 6258R CPU @ 2.70GHz 28 cores each (total of 56 cores) and 512G of RAM. However, for training, we used at most 8 cores.
	
	\subsection{Output}
	\label{sec:output}
	
	At the end of training, we have an explanation in DNF form for each Qcode. We use these DNF rules for classifying the new data. We check the message against each explanation and output the list of Qcodes for which the message evaluates to TRUE in the corresponding DNF formula. This is our candidate list for that message. As described in Section~\ref{sec:weights}, we  prioritize the Qcodes in the candidate list using candidate weights. We then select the top $K$ ($1 \leq K \leq 10$) candidates (Qcodes) from the list and discard the other candidates. \kp{The resulting candidate list, along with the candidate weights, is passed to a human (domain expert) to select the final QCode.}
	
	To evaluate the performance of our approach, we measure the accuracy and the average list length for each reduced candidate list and also for each complete candidate list.

	\section{Computational Results}
	\label{sec:largedata}
	
	This section presents the computational results obtained on the real data provided by our industrial partner Thales (see Section \ref{sec:preprocess}).
	
	\subsection{Tuning experiments}
	
	First, Table \ref{tab:tuning-large} reports experimental results obtained from complete target lists when varying the values of the  hyperparameters listed in the first four columns. For each hyperparameter configuration, we record the total training time (in seconds), accuracy, maximum candidate list length and average candidate list length per Qcode. For each example, if the candidate list contains the actual Qcode, we count it as a successful prediction. The accuracy is measured by the number of successful predictions divided by the total number of examples evaluated. As seen in the Table \ref{tab:tuning-large} first six rows, it is easy to achieve high accuracy with very high candidate list length. The accuracy drops as the parameters make the algorithm more selective to reduce the candidate list length. Also note that, while the average candidate list length in the last three rows of Table \ref{tab:tuning-large} is small enough to be interpreted by humans, the maximum candidate list length is very high. To overcome this, we need to reduce the list length by removing the candidates with lower weights $W_{diff}$.
	
	\begin{table}[H]
		\tbl{Hyperparameter tuning experimental results}{
			\begin{tabular}{ccccccccccc}
				\toprule
				& & & & Training time & \multicolumn{3}{c}{Training} & \multicolumn{3}{c}{Validation} \\ \cmidrule{6-8}\cmidrule{9-11}
				$P$ & $C$ & $I$ & $N$  & (seconds) & Accuracy & \multicolumn{2}{c}{List length} & Accuracy & \multicolumn{2}{c}{List length} \\ \cmidrule{7-8}\cmidrule{10-11}
				& & & & & & Max & Avg &  & Max & Avg  \\ \midrule
				2 & 10 & 10 & 2 & 916 & 0.96  & 72 & 9.93 & 0.92 & 72 & 9.97 \\ 
				2 & 10 & 20 & 2 & 988 & 0.96 & 73 & 10.21 & 0.92 & 70 & 10.28 \\ 
				2 & 10 & 30 & 2 & 993 & 0.96 & 71 & 10.03 & 0.92 & 71 & 10.02 \\ 
				3 & 10 & 30 & 2 & 1025 & 0.97 & 78 & 11.35 & 0.93 & 75 & 11.35 \\ 
				4 & 10 & 30 & 2 & 1035 & 0.98  & 78 & 12.22 & 0.94 & 78 & 12.29 \\ 
				5 & 10 & 30 & 2 & 1028 & 0.98  & 80 & 12.48 & 0.94 & 77 & 12.56 \\ 
				4 & 10 & 30 & 5 & 1297 & 0.95 & 60 & 7.53 & 0.91  & 60 & 7.57 \\ 
				4 & 10 & 30 & 10 & 2006 & 0.93 & 50 & 5.15 & 0.89 & 49 & 5.14\\ 
				4 & 10 & 30 & 20 & 3356 & 0.91 & 46 & 3.47 & 0.86 & 46 & 3.48 \\ 
				4 & 20 & 30 & 20 & 4488 & 0.94 & 46 & 3.15 & 0.87 & 46 & 3.14 \\ 
				4 & 30 & 30 & 20 & 5381 & 0.95 & 47 & 3.05 & 0.87 & 46 &3.02 \\ \bottomrule
				
		\end{tabular}}
		\label{tab:tuning-large}
	\end{table}
	
	Table \ref{tab:training-stats} shows various statistics related to training for various parameter configurations (identified by the first four columns). We can see that most of the time is used for solving the SP using the CP-SAT solver. Compared to that, the time required to solve the RMP and the integer RMP is very small. We see from the last three rows that the average clause length per Qcode increases with $C$ as expected. In the fifth column of Table \ref{tab:training-stats} ('Number of MPs solved'), we record the number of times (out of 204 Qcodes) for which the MP is solved, i.e., where the subproblem cannot find any column with negative reduced cost within a tolerance of $10^{-2}$. We also record the number of columns added by the heuristic (described in Section~\ref{sec:solving}) and by CP-SAT. This helps us measure the effectiveness of the heuristic. Table \ref{tab:training-stats} also shows the average clause length per Qcode and the time taken by various parts of our column generation approach. 
	
	\setcounter{table}{3}
	\begin{sidewaystable}
		\tbl{Hyperparameter tuning training statistics}
		{\begin{tabular}{cccc>{\centering}p{0.1\textwidth}cc>{\centering}p{0.1\textwidth}ccccc} \toprule
				
				& & & & Number of MPs solved & \multicolumn{2}{>{\centering}p{0.2\textwidth}}{Average number of columns added per Qcode} & Average clause length per Qcode & \multicolumn{5}{c}{Average Time per Qcode (in seconds)} \\ \cmidrule{9-13}\cmidrule{6-7}
				$P$ & $C$ & $I$ & $N$ & & Heuristic & CP-SAT & &  RMP & CP-SAT & Heuristic & Integer RMP & Total \\ \midrule
				2 & 10 & 10 & 2 & 181 & 14.90 & 1.58 & 1.37 & 0.003 & 3.44 & 1.00 & 0.01 & 4.49 \\ 
				2 & 10 & 20 & 2 & 203 & 15.11 & 1.64 & 1.37 & 0.003 & 3.63 & 1.17 & 0.01 & 4.84 \\ 
				2 & 10 & 30 & 2 & 204 & 14.80 & 1.74 & 1.35 & 0.003 & 3.74 & 1.08 & 0.01 & 4.87 \\ 
				3 & 10 & 30 & 2 & 204 & 16.06 & 2.08 & 1.37 & 0.003 & 3.88 & 1.09 & 0.01 & 5.03 \\ 
				4 & 10 & 30 & 2 & 204 & 15.84 & 1.86 & 1.37 & 0.003 & 3.93 & 1.10 & 0.01 & 5.07 \\ 
				5 & 10 & 30 & 2 & 204 & 15.71 & 2.23 & 1.36 & 0.003 & 3.87 & 1.12 & 0.01 & 5.04 \\ 
				4 & 10 & 30 & 5 & 204 & 13.71 & 3.56 & 1.49 & 0.003 & 5.51 & 0.78 & 0.01 & 6.36 \\ 
				4 & 10 & 30 & 10 & 204 & 10.36 & 5.74 & 1.55 & 0.004 & 9.18 & 0.55 & 0.02 & 9.84  \\ 
				4 & 10 & 30 & 20 & 204 & 5.74 & 8.97 & 1.66 & 0.004 & 16.04 & 0.24 & 0.02 & 16.45 \\ 
				4 & 20 & 30 & 20 & 204 & 5.96 & 12.23 & 1.88 & 0.005 & 21.48 & 0.32 & 0.02 & 22.00 \\ 
				4 & 30 & 30 & 20 & 203 & 5.80 & 13.80 & 1.95 & 0.006 & 25.81 & 0.36 & 0.02 & 26.38 \\ \bottomrule
		\end{tabular}}
		\label{tab:training-stats}
	\end{sidewaystable}
	
	From the results in Tables \ref{tab:tuning-large} and~\ref{tab:training-stats}, the effect of increasing each hyperparameter value can be summarized as follows.
	
	\begin{enumerate}
		\item[$P$:] Improves accuracy and increases list lengths.
		\item[$C$:] Improves accuracy, reduces list lengths, and increases average clause length.
		\item[$I$:] For more Qcodes, the MPs are solved before reaching the maximum number of iterations. We observe that for our tests a maximum of $I = 30$ column generation iterations is sufficient to solve the MP for at least 203 of the 204 Qcodes.
		However, there is no clear signal on accuracy and list lengths. Increases solving time (see the first three rows of Table \ref{tab:tuning-large})  
		\item[$N$:] Reduces accuracy, reduces list length, increases average clause length, and increases solving time.
	\end{enumerate}

	Second, Table \ref{tab:red-lists} presents the results for reduced target lists, by considering different maximum length values $K \in \{1,\dots, 10\}$. For this experiment, we fix what turns out to be the best configuration of parameters, namely, $P=4$, $C=30$, $I=30$, and $N=20$. 
	
	\begin{table}[H]
		\tbl{Accuracy for the hyperparameter configuration ($P=4,C=30,I=30,N=20$) by using different candidate list lengths $K$. The results are compared with the last line of Table \ref{tab:tuning-large} denoted as $K = \infty$.}{
			
			\begin{tabular}{ccc}
				\toprule
				$K$ & Training & Validation \\ \midrule
				1 & 0.65 & 0.60 \\ 
				2 &  0.86& 0.80 \\ 
				3 & 0.90 & 0.84 \\ 
				4 & 0.92 & 0.85 \\ 
				5 & 0.93 & 0.86 \\ 
				6 & 0.94 & 0.86 \\ 
				7 & 0.94 & 0.87 \\ 
				8 & 0.94 & 0.87 \\ 
				9 & 0.95 & 0.87 \\ 
				10 & 0.95 &0.87 \\ \midrule 
				$\infty$ & 0.95 & 0.87\\ \bottomrule
		\end{tabular}}
		\label{tab:red-lists}
	\end{table}
	
	From the results in Table \ref{tab:red-lists}, we can observe that we do not achieve good accuracy for very small values of $K$, but the candidate lists with $K \geq 3$ produces good accuracy values, while $K=9$ already matches the accuracy in Table \ref{tab:tuning-large}, where $K$ is not limited. It will be shown in Table \ref{tab:ml-models-large} that already $K = 3$ gives results that are comparable to other machine learning algorithms.
	
	Finally, for the best configuration of parameters, i.e., $P=4$, $C=30$, $I=30$, and $N=20$, we summarize the results in Table \ref{tab:best-res}, including the test accuracy with and without limited target lists. The first column shows the number of examples that had more than one Qcodes in the candidate list. 
	Furthermore, the second part of the table considers the case in which we restrict the evaluation to the Qcodes with at least 50 examples in the training data. As expected, we observe higher accuracy and lower candidate list lengths with the restricted evaluation. This shows that using more data for each Qcode can improve the accuracy and reduce the candidate list lengths.
	
	\begin{table}[H]
		\tbl{Results for best hyperparameter values. The last three lines consider only the Qcodes with at least 50 examples.}{
			
			\begin{tabular}{c>{\centering}p{0.1\textwidth}>{\centering}p{0.1\textwidth}cc}
				\toprule
				Dataset  & Number of NOTAMs with more than one Qcode in the list & Average list length & Accuracy & Accuracy ($K=4$) \\ \midrule
				Training & 8335 & 3.05 & 0.95 & 0.92 \\ 
				Validation & 2749 & 3.02  & 0.87 & 0.85 \\ 
				Test & 2760 & 3.01 & 0.87 & 0.86 \\  \midrule
				Training & 3758 & 1.49 & 0.94 & 0.94 \\ 
				Validation & 1235 & 1.46  & 0.90 & 0.89  \\ 
				Test & 1206 & 1.44 & 0.90 & 0.89 \\ \bottomrule

		\end{tabular}}
		\label{tab:best-res}
	\end{table}
	
			%
	
	As an example, we present the boolean decision rule explanation generated for the Qcode 'LAAS':  
	\begin{itemize}
		\item[.] ['U/S' AND 'ALS'] OR 
		\item[.] ['RWY' AND 'U/S' AND 'MAINT'] OR 
		\item[.] ['U/S' AND 'SYSTEM' AND 'APCH'] OR 
		\item[.] ['U/S' AND '24'] OR 
		\item[.] ['U/S' AND 'MAINT' AND '02'] OR 
		\item[.] ['RWY' AND 'U/S' AND '19'].
	\end{itemize} 
	The first part of the Qcode 'LA' indicates 'Approach lighting system (specify runway and type)' and the second part of the Qcode 'AS' means 'Unserviceable'. The positive and negative accuracies for the clauses are shown in Table \ref{tab:sample-rule-weights}. For this Qcode, we used 3003 data points for training. We can see that the last four clauses are only selected to cover a few positive training data points. This is where the model tries to overfit the training data. We see that the weights for these clauses are very low compared to the other two clauses. 
	
	\begin{table}[H]
		\tbl{Weights for clauses for the boolean decision rule for label 'LAAS'}{
			
			\begin{tabular}{cccc}
				\toprule
				Clause & Positive accuracy & Negative accuracy & $W_{diff}^*$ \\ \midrule
				'U/S' AND 'ALS' & 0.937 & 0.006 & 2795.79   \\ 
				'RWY' AND 'U/S' AND 'MAINT' & 0.035  & 0.002 & 99.10  \\ 
				'U/S' AND 'SYSTEM' AND 'APCH' & 0.007 & 0.000 & 21.02   \\ 
				'U/S' AND '24' & 0.007  & 0.000 & 21.02  \\ 
				'U/S' AND 'MAINT' AND '02' & 0.007  & 0.000 & 21.02  \\ 
				'RWY' AND 'U/S' AND '19' & 0.007  & 0.000 & 21.02  \\ \bottomrule
				
		\end{tabular}}
		\label{tab:sample-rule-weights}
	\end{table}
	
	\subsection{Comparison with standard ML models}
	
	In this section, we compare the accuracy of our approach with those that can be achieved by using \kp{the following} standard machine learning algorithms.
	
	\kp{
		\begin{enumerate}
			\item Max Class Classifier: Picks the most frequent class in the training data as the prediction. This is typically used for benchmarking.
			\item Bernoulli Naive Bayes: This classifier assumes that the data comes from a Bernoulli distribution. It uses the training data to find the parameters for the underlying Bernoulli distribution. Finally, it uses Bayes' rule to estimate the probability of an example belonging to a fixed class.
			\item Decision trees: This classifier builds a tree. Each internal node of this tree has a split check. The data point follows the branch corresponding to the split check decision. Finally, each leaf node is assigned a class. The data point is classified into the class of the leaf node it reaches. This is the most interpretable classifier among the ones compared.
			\item XGBoost: This is a combination of multiple simple classifiers (like decision trees) that are trained sequentially.
			\item Linear SVM: This is a binary classifier that separates the datapoints of the two classes by a line in such a way that the distance of the nearest datapoint from the line is maximized. This is the only classifier that is designed for binary classification and extended to multiclass classification.
			\item ReLU neural network: ReLU is a piecewise linear function defined as $ReLU(x) = max(0,x)$. The neural network passes the input through a series of non linear functions ($ReLU(wx + b)$) in order to predict the final class. The goal is to be able to learn a much more complex function. 
		\end{enumerate}
	}
	
	\kp{The hyperparameters for all the models except XGBoost were tuned using Scikit-learn's GridsearchCV method. For XGBoost, we used Scikit-learn's RandomizedSearchCV method because of the higher training time. The values that achieved the highest accuracy on the validation set were selected. We report the values considered for each hyperparameter in Table \ref{tab:ml-models-tuning}.}
	
	\begin{table}[h]
		\tbl{\kp{Hyperparameter tuning for the standard ML algorithms.}}{
			\begin{tabular}{p{0.25\textwidth}p{0.25\textwidth}p{0.25\textwidth}p{0.15\textwidth}}
				\toprule
				\kp{Model} & \kp{Hyperparameter (Scikit-learn)} &\kp{Values considered}& \kp{Selected value}\\
				\midrule
				\kp{Bernoulli Naive Bayes} & \kp{$\alpha$} & \kp{0.01, 0.02, 0.03, 0.04, 0.05, 0.06, 0.07, 0.08, 0.09, 0.1} & \kp{0.04}\\
				\kp{Decision trees} & \kp{criterion} & \kp{gini, entropy} & \kp{gini}\\
				& \kp{max depth} & \kp{4,   8,  16,  32,  64, 128} & \kp{64}\\
				& \kp{max features} & \kp{sqrt, log2} & \kp{sqrt}\\
				& \kp{max leaf nodes} & \kp{   4,    8,   16,   32,   64,  128,  256,  512, 1024, 2048} & \kp{2048}\\
				& \kp{min samples leaf} & \kp{1, 11, 21, 31, 41, 51} & \kp{1}\\
				& \kp{min samples split} & \kp{2, 12, 22, 32, 42} & \kp{12}\\
				& \kp{splitter} & \kp{best, random} & \kp{random}\\
				
				\kp{XGBoost} & \kp{colsample bytree} & \kp{All reals in (0.7, 1.0)} & \kp{0.75}\\
				& \kp{gamma} & \kp{All reals in (0,0.5)} & \kp{0.179}\\
				& \kp{learning rate} & \kp{All reals in (0.03, 0.33)} & \kp{0.255}\\
				& \kp{max depth} & \kp{All integers in [2 ,6]} & \kp{5}\\
				& \kp{n estimators} & \kp{All integers in [100, 150]} & \kp{112}\\
				& \kp{subsample} & \kp{All reals in (0.6, 1.0)} & \kp{0.828}\\
				& \kp{tree method} & \kp{hist} & \kp{hist}\\
				& \kp{objective} & \kp{multi:softprob} & \kp{multi:softprob}\\
				
				\kp{Linear SVM} & \kp{C} & \kp{0.01, 0.1, 1.0, 10.0, 100.0} & \kp{1.0}\\
				& \kp{max iter} & \kp{10000000} & \kp{10000000}\\
				\bottomrule
		\end{tabular}}
		\label{tab:ml-models-tuning}
	\end{table}
	
	We used the python Scikit-learn library for evaluating the results that were generated on Google Colab (free version) and are reported in Table \ref{tab:ml-models-large}, \kp{where $K$ denotes the candidate list length}. The last three rows in this table provide the results of the proposed Boolean decision rule approach, using the best hyperparameter configuration \kp{($P = 4, C = 30, I = 30, N = 20$)} for three maximum list lengths $K = 3, 4, 5$, respectively.
	
	\begin{table}[t]
		\tbl{Results of different ML models for multiclass classification.}{
			\begin{tabular}{p{0.25\textwidth}cccc}
				\toprule
				Model &\kp{$K$}& \multicolumn{3}{c}{Accuracy}\\ \cmidrule{3-5}
				& & Training & Validation & Test \\ \midrule
				Max Class Classifier  &\kp{1} & 0.06 & 0.06 & 0.06\\*[3pt] 
				Bernoulli Naive Bayes  &\kp{1} & 0.84 & 0.77 & 0.78 \\*[3pt]
				Decision \kp{t}rees  &\kp{1} & 0.84 & 0.76 & 0.75 \\*[3pt] 
				XGBoost &\kp{1}& \kp{0.94} & \kp{0.85} & \kp{0.85}\\
				&\kp{3}& \kp{0.97} & \kp{0.89} & \kp{0.89}\\
				&\kp{4}& \kp{0.97} & \kp{0.89} & \kp{0.89}\\
				&\kp{5}& \kp{0.97} & \kp{0.89} & \kp{0.89}\\*[3pt]
				
				Linear SVM &\kp{1} & 0.94 & 0.84 & 0.84\\ 
				&\kp{3} & \kp{0.95} & \kp{0.86} & \kp{0.86}\\
				&\kp{4} & \kp{0.95} & \kp{0.86} & \kp{0.86}\\
				&\kp{5} & \kp{0.95} & \kp{0.86} & \kp{0.86}\\*[3pt]
				ReLU Neural Network &\kp{1} & 0.95 & 0.84 & 0.84\\
				&\kp{3} & \kp{0.98} & \kp{0.88} & \kp{0.88}\\ 
				&\kp{4} & \kp{0.98} & \kp{0.88} & \kp{0.88}\\ 
				&\kp{5} & {\kp{0.99}} & \kp{0.88} & \kp{0.88}\\*[3pt] 
				Boolean Decision Rule &\kp{3} & 0.90 & 0.84 & 0.84 \\ 
				&\kp{4} & 0.92 & 0.85 & {0.86}\\ 
				&\kp{5} & 0.93 & {0.86} & {0.86}\\ \bottomrule
		\end{tabular}}
		\label{tab:ml-models-large}
	\end{table}

	The results in Table \ref{tab:ml-models-large} show that \kp{XGBoost} gives the best accuracy in the test data \kp{followed by a (relatively) simple 5-layer ReLU neural network and a Linear SVM for single target prediction ($K=1$). For these} models, the best accuracy on the test data is comparable to the accuracy achieved on validation and test data by our column generation approach. However, \kp{d}ecision trees, which are more interpretable than SVM and neural networks, cannot match the accuracy achieved by our proposed prediction, thus corroborating it. \kp{For a fair comparison, we also report the accuracies for XGBoost, Linear SVM, and 5-layer ReLU neural network with top $K(>1)$ candidate predictions instead of a single QCode prediction. These models output a prediction probability for each class. We can extract the top $K$ predictions using these probabilities. In this setting, the column generation approach is still comparable to Linear SVM in terms of accuracies. However, XGBoost and neural network are significantly better than the column generation approach, but they lack interpretability, the main goal of our investigation.}
	
	\kp{We did not consider larger neural networks for the comparision. However, Thales reported that they could get 93.9\% accuracy (without explanations) using BERT \citep{bert19}, a more advanced machine learning algorithm for text classification and a lot more data (about 100,000 examples).}

	\subsection{Practical Impact}
	
	The column generation algorithm is not able to match \kp{the BERT} accuracy result, but it provides an explanation for every NOTAM classification.  One of the major issues is, however, the scalability of the algorithm. We note that some of the algorithms for Boolean decision rule generation that are based on heuristics \citep[for example, RIPPER from][]{cohen1995fast} require less training time. An interesting venue for future development is to use such methods to generate the starting columns for the MP \eqref{origmodel}.
	
	From an operational perspective, it is crucial for Thales to capture explanations for NOTAM classification, as explanations do contain information essential to personnel concerned with flight operations. Prior to the current work, this was not accessible in any way, while communicating explanations of NOTAM classification is now possible, and useful for validating NOTAM categories. The loss in accuracy has been strongly balanced by the explanation, which in turn is used for validation, providing very useful insight for human to trust the classification results, and strongly support flight operation.
	
	\section{Conclusion and Future Work}
	\label{sec:conclusion}
	
	
	In this paper, we extended the work of \cite{Dash2018} to a real-world multiclass classification problem. We could generate interpretable boolean decision rules for classifying NOTAMs into Qcodes. In the process, we demonstrated a method to tackle the issue of multiple candidate prediction that results from one-vs-rest extension of the binary classification algorithm in \cite{Dash2018}. The accuracy we achieved using this method were comparable to some of the state-of-the-art machine learning algorithms like SVM and small neural networks.
	
	We showed a method to generate candidate weights that help us find more relevant candidates when more than one candidate is predicted. Although the way we generate the candidate weights seems reasonable, it does not necessarily give the best weights, as seen in Table \ref{tab:red-lists}. We plan to work on finding improved ways to compute the candidate weights to get even better accuracy for smaller candidate lists, keeping in mind that the method for computing the weights should be easy to interpret by humans.
	
	Finally, we also want to explore the effect of using the complete branch-and-price algorithm to generate the explanation instead of only generating columns in the root node. This will obviously result in longer training times, but can potentially help us find better Boolean decision rules. Such rules might then be studied to understand the underlying limitations of classifying NOTAMs using Boolean decision rules.
	
	\section*{Acknowledgements}
	The first author is supported by a research grant from the project DEpendable and Explainable Learning (DEEL). We would like to thank Giuliano Antoniol for coordinating between DEEL and Polytechnique Montreal. \kp{We thank two anonymous reviewers for their detailed reading and useful comments.}
	
	\section*{Declaration of Interest} 
	In accordance with Taylor \& Francis policy and his ethical obligation as a researcher, the fourth author (F. Lecue) is reporting that he has a financial and/or business interests in a company that may be affected by the research reported in the enclosed paper. He has disclosed those interests fully to Taylor \& Francis, and he has in place an approved plan for managing any potential conflicts arising from that involvement. The other authors declare that they have no competing interest. 
	
	\section*{Data availability statement} 
	After approval from Thales, all data used for our experiments should be made available if the paper is accepted for publication. 
	
	
	
	
	
	\bibliographystyle{apacite}
	\bibliography{references}
	
\end{document}